\documentclass[11pt]{article}
\usepackage{acl}
\usepackage{fontspec,xeCJK}
\IfFontExistsTF{Songti SC}{
  \setCJKmainfont{Songti SC}
  \setCJKsansfont{Heiti SC}
  \setCJKmonofont{Songti SC}
}{
  \setCJKmainfont{FandolSong-Regular.otf}[BoldFont=FandolSong-Bold.otf,ItalicFont=FandolKai-Regular.otf]
  \setCJKsansfont{FandolHei-Regular.otf}[BoldFont=FandolHei-Bold.otf]
  \setCJKmonofont{FandolFang-Regular.otf}
}
\usepackage{latexsym,microtype,etoolbox}
\usepackage{amsmath,amssymb,booktabs,tabularx,array,graphicx}
\usepackage{colortbl}
\usepackage{placeins}
\usepackage{algorithm}
\usepackage[noend]{algpseudocode}
\algrenewcommand{\algorithmicrequire}{\textbf{Input:}}
\algrenewcommand{\algorithmicensure}{\textbf{Output:}}
\renewcommand{\abstractname}{Abstract}

\newcommand{\bench}{TandemBench}
\newcommand{\ind}{\mathbb{I}}

\newcommand{\hfmodelhref}[2]{\href{#1}{\raisebox{-0.12ex}{\includegraphics[height=0.9em]{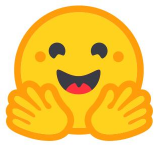}}\hspace{0.25em}\textcolor{black}{#2}}}
\title{TandemBench: Benchmarking Decision Models\\for Multi-Agent Collaboration}
\author{
  Baohang Li\textsuperscript{1} \quad
  Xiaocheng Feng\textsuperscript{1,2} \quad
  Bing Qin\textsuperscript{1} \\
  \textsuperscript{1}Harbin Institute of Technology \qquad
  \textsuperscript{2}Peng Cheng Laboratory \\
  \texttt{baohangli@ir.hit.edu.cn}
}
\begin{document}
\maketitle
\renewcommand{\abstractname}{Abstract}
\begin{abstract}
Decision models choose among predefined options in tasks such as routing and action selection. In multi-agent collaboration, a decision determines how members divide work, exchange resources, and act in parallel. We introduce \bench{}, a benchmark of 1,200 questions in multi-agent cooking scenarios where coordination choices affect goal completion. Models receive complete global information and candidate team plans, with reference answers established through execution and exact search. Models select an optimal team plan and, in paired variants that retain or remove it, judge whether better arrangements exist beyond the supplied options. We evaluate 40 native decision models and compare them with 10 general-purpose language models. The best native decision model achieves 62.50\% decision accuracy. Native decision models remain close to chance in optimality verification. Paired tests reveal that some report better alternatives regardless of whether an optimal plan is already present. These findings show that team-plan decision accuracy alone can conceal systematic errors in judging whether the supplied options contain an optimal arrangement. The dataset and evaluation code are publicly available at \url{https://github.com/lililibaohang/TandemBench}.
\end{abstract}

\section{Introduction}

\begin{figure}[!t]
\centering
\includegraphics[width=\columnwidth]{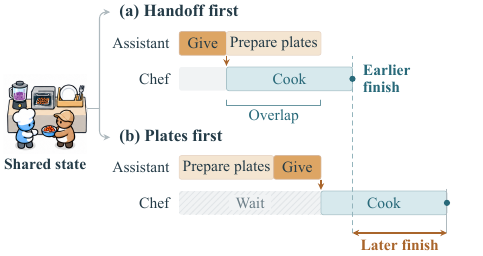}
\caption{One member's action order affects the whole team. (a) Handing over ingredients first enables parallel cooking and plate preparation. (b) Preparing plates first leaves the chef waiting.}
\label{fig:example}
\end{figure}

Decision models choose among predefined options in tasks such as routing and action selection~\citep{duan2026jeval,geuter2026opendecisions}. In multi-agent collaboration, a single decision can determine how team members coordinate their actions toward a shared goal. For example, an assistant can supply an ingredient before preparing plates, allowing the chef to cook while the assistant continues preparation. Preparing plates first delays the handoff and leaves the chef waiting (Figure~\ref{fig:example}). Judging a team plan therefore requires considering how members' actions depend on one another and jointly contribute to the goal.

Existing decision benchmarks examine accuracy, calibration, and robustness~\citep{duan2026jeval,zhuang2026decisionpivot,lan2026decisionrobust}, while AgentDecisionBench evaluates tool and action choices in agent workflows~\citep{zhu2026jevagentharness}. We focus on decisions that specify joint plans for multiple team members. Several plans may be feasible, yet their coordination choices can lead to different outcomes for the team. Evaluating these choices requires a reference for the best achievable outcome from the given state, since even the best supplied plan may be suboptimal. We therefore assess whether models select an optimal team plan and recognize when better arrangements exist beyond the supplied options.

We introduce \bench{}, comprising 1,200 questions from 200 scenarios across 30 recipe tasks in Collab-Overcooked~\citep{sun2025collab}. We select states where action ordering or resource handoffs affect task completion even when the remaining actions are optimized. Each question provides complete global information and a candidate set containing multiple feasible team plans. Execution and exact search establish candidate outcomes and the best achievable result from each state. Paired variants retain or remove the optimal plan while keeping the state and goal fixed, allowing us to distinguish correct reports of better alternatives from false alarms.

\begin{figure}[!t]
\centering
\includegraphics[width=\columnwidth]{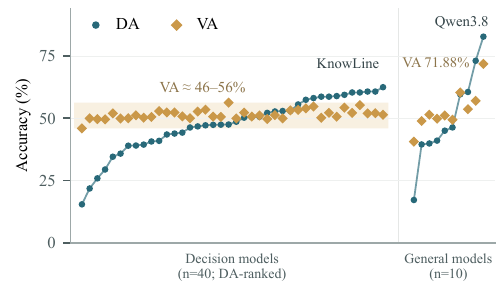}
\caption{Decision accuracy (DA) and verification accuracy (VA), with models sorted by increasing DA within each family.}
\label{fig:headline}
\end{figure}

We evaluate 40 native decision models and compare them with 10 general-purpose language models. KnowLine leads the native models in decision accuracy (DA) at 62.50\% (Figure~\ref{fig:headline}; Tables~\ref{tab:results} and~\ref{tab:general-results}), while their verification accuracy (VA) remains near chance (50\%). Paired tests reveal two contrasting errors: some models report a better plan whether or not the optimum is supplied, while others select a supplied plan even when it is suboptimal. Thus, DA alone misses systematic errors in optimality judgments.

Our contributions are as follows.
\begin{itemize}
    \item We introduce \bench{}, a 1,200-question benchmark for evaluating decision models on how multiple agents coordinate their actions toward a shared goal, given complete global information.
    \item We ground comparisons among candidate coordination plans in execution and exact search, and use paired tests to assess whether models recognize when the supplied options omit the optimal arrangement.
    \item We show that decision accuracy alone misses systematic errors in optimality verification, and analyze how these decisions vary across short- and long-horizon goals and option orders.
\end{itemize}

\section{Related Work}
\label{sec:related}

\paragraph{Decision Models.}
Decision-model evaluation covers both individual choices and decisions within agent workflows. JEVal and OpenDecisions assess accuracy, calibration, and inference efficiency~\citep{duan2026jeval,geuter2026opendecisions}, while DecisionRobust and DecisionPivot examine adversarial robustness and controlled input changes, respectively~\citep{lan2026decisionrobust,zhuang2026decisionpivot}. AgentBench and AgentBoard evaluate sequential decisions through task success and progress~\citep{liu2024agentbench,ma2024agentboard}; AgentDecisionBench spans isolated choices, recorded agent states, and online control~\citep{zhu2026jevagentharness}. TandemBench takes a joint team plan as the decision unit, so one choice specifies actions for multiple members toward a shared goal.

\paragraph{Multi-Agent Collaboration.}
CoELA and RoCo study communication and coordinated execution, while MultiAgentBench and PARTNR evaluate collaboration in interactive settings~\citep{zhang2024coela,mandi2024roco,zhu2025multiagentbench,chang2025partnr}. CoordQA and RoCoBench-Text assess coordination choices and reasoning through static questions~\citep{agashe2025coordination,mandi2024roco}. ParaCook supplies CP-SAT optimal references for abstract scheduling~\citep{zhang2026paracook}, while CoCoBench evaluates coordination against optimal makespan under a skill abstraction~\citep{chen2026cocobench}. TandemBench builds on Collab-Overcooked's asymmetric cooking environment~\citep{sun2025collab}. It evaluates a single model's choices among executable team plans under complete global information, using an exact optimality reference from the same state.

\paragraph{Planning and Verification.}
Planning benchmarks assess plan generation and reasoning about actions, state transitions, and constraints~\citep{valmeekam2023planbench,kokel2025acpbench,xie2024travelplanner,zheng2024naturalplan}. Plan-optimality verification asks whether a given plan is optimal~\citep{lin2024planoptimality}, while optimality certificates provide independently checkable evidence~\citep{dold2025prooflogging}. DecisionPivot uses controlled contrasts to examine how decisions respond to input changes~\citep{zhuang2026decisionpivot}. TandemBench asks whether any supplied joint team plan attains the state optimum. Paired variants retain or remove the optimal plan while keeping the state and goal fixed, testing whether model judgments track its presence.

\section{Multi-Agent Collaborative Decision-Making}
\label{sec:task}
\subsection{Task Definition}
A single decision model receives a complete global state, the environment rules, a shared goal, and candidate team plans. Each candidate specifies an action program for each of two team members. The global state records both members' configurations, task-relevant resources, and ongoing processes.

The two programs execute concurrently under the environment rules. Their dependencies and resource handoffs determine when each member can proceed (Figure~\ref{fig:example}). A candidate may cover the full task or prescribe a stage that ends before the final goal. Appendix~\ref{app:semantics} specifies the execution semantics.

\subsection{Decision Objectives}
We evaluate a team plan against either an intermediate milestone or the final task goal, referred to as short- and long-horizon goals. Completion cost is measured in joint execution steps.

\paragraph{Short-horizon goals.}
We count the steps until the specified intermediate milestone is first reached.

\paragraph{Long-horizon goals.}
We count the steps until the final goal is reached. If the prescribed stage ends before completion, the cost is its executed steps plus the minimum remaining steps from the resulting state.

Both goal settings apply to team-plan decisions and optimality verification. Short- and long-horizon questions from the same state use their respective execution protocols and may have different candidate programs. Appendix~\ref{app:formal-decisions} provides the formal definitions.

\begin{figure*}[!t]
\centering
\includegraphics[width=\textwidth]{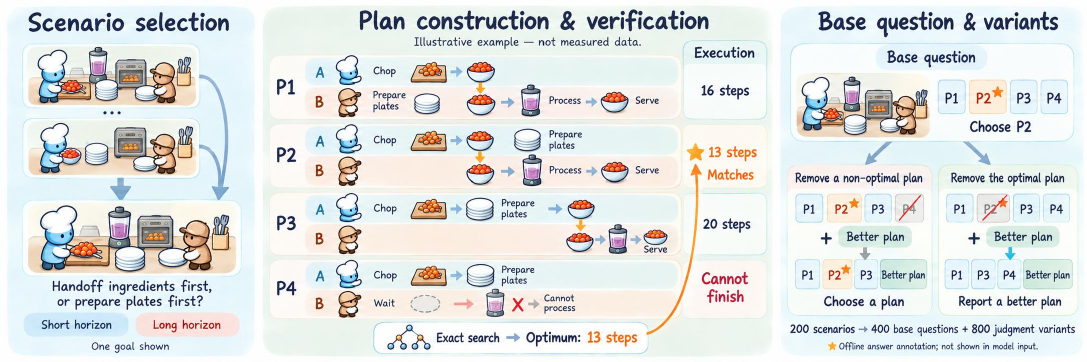}
\caption{Construction of \bench{}. We select scenarios with consequential coordination choices and construct candidate team plans. Execution and exact search establish reference answers. Retaining or removing the optimal plan yields paired verification questions.}
\label{fig:construction-pipeline}
\end{figure*}

\subsection{Optimal Decisions}
Both tasks use the state optimum: the minimum completion cost of any legal joint execution from the given state under the specified goal and action domain.

For \emph{team-plan decision-making}, the model chooses among four candidate plans. Exactly one attains the state optimum and is the correct answer.

Plan-optimality verification asks whether a given plan is optimal~\citep{lin2024planoptimality}. We evaluate whether the supplied set contains such a plan. Each verification question includes three concrete plans and a fourth option stating that a feasible plan can reach the goal in fewer joint steps than all three. This statement is true exactly when none of the supplied plans attains the state optimum.

The two tasks are scored separately. Decision accuracy scores the optimal concrete choice. Verification accuracy scores whether the better-plan statement is correctly accepted or rejected; any valid concrete choice counts as rejection. Section~\ref{sec:evaluation} defines the metrics, and Appendix~\ref{app:formal-decisions} gives the formal decision criteria.

\section{TandemBench}
\label{sec:construction}
We construct \bench{} from states where coordination choices affect task completion. Alternative team plans make these choices explicit, and execution and exact search establish their outcomes and the state optimum. Each verified base question yields paired variants that retain or remove the optimal plan (Figure~\ref{fig:construction-pipeline}).

\subsection{Environment and Tasks}
We use 30 recipe tasks from Collab-Overcooked~\citep{sun2025collab}, where a chef and an assistant work in separate kitchen regions with asymmetric access to ingredients, appliances, and counters. Access restrictions make some actions depend on resources supplied by the other member, while recipe dependencies constrain action order. Automatic processing allows preparation to overlap with cooking (Figure~\ref{fig:example}). Carrying limits and shared-counter availability constrain handoff locations and timing. Appendix~\ref{app:semantics} gives the execution rules.

\subsection{Scenario Selection}
\label{sec:scenario-selection}
We select states where fixing a coordination choice changes the best achievable completion cost. Candidate states come from verified optimal delivery trajectories, at events that change available materials, processing status, or handoff opportunities. Execution prefixes establish reachability.

At each state, we compare alternative orders of two required events performed by the same member or alternative locations for a required handoff. Exact search determines the minimum delivery cost subject to each choice, optimizing all remaining actions. A state qualifies when two feasible choices are established: one attains the state optimum, while the other requires at least three additional joint steps.

We select 200 qualifying scenarios across all 30 tasks, prioritizing coverage of event orderings, handoff directions, comparison families, and material stages. Screening uses the final delivery goal and constraints on individual coordination choices. Concrete candidate programs are constructed and evaluated separately for the short- and long-horizon questions. Algorithm~\ref{alg:scenario-selection} in Appendix~\ref{app:construction-details} details screening and coverage.

\subsection{Instance Construction}
\label{sec:instance-construction}
Each scenario supplies a short-horizon and a long-horizon base question. From each verified base question, we derive a pair of optimality-verification variants.

\subsubsection{Question Construction}
\label{sec:option-construction}
Candidate plans express different ways to coordinate the same team toward the specified goal. They vary in work order, handoff location, preparation scope, work overlap, or waiting conditions. For example, changing the order of ingredient supply and utensil preparation affects when processing can begin and which work can overlap. Handoff choices also depend on members' positions and counter availability. Appendix~\ref{app:example} gives concrete candidate descriptions and costs.

Each base question contains four plans, with at least three feasible and exactly one attaining the state optimum. Multiple feasible alternatives require comparing arrangements that can complete the goal. We aim to include alternatives with small cost gaps when they express distinct coordination choices. Candidate programs may differ between the short- and long-horizon objectives.

To construct a verification pair, we remove the optimal plan in one variant and a predetermined nonoptimal plan in the other. Both add the same statement that a feasible plan can reach the goal in fewer joint steps than all three remaining plans. The state, goal, execution rules, and retained programs remain unchanged. The resulting four-option pair supports comparison of correct reports when the optimum is missing and false alarms when it is present.

\subsubsection{Answer Annotation}
\label{sec:quality-control}
Candidate execution records the steps taken, goal attainment, and resulting state. Exact search establishes the state optimum and any optimal continuation needed for long-horizon evaluation. Together, these determine candidate costs under the objectives in Section~\ref{sec:task}. Matching candidate costs to the state optimum identifies the correct base answer. In the verification pair, the better-plan statement is false when the optimum is retained and true when it is removed.

Noncompletion requires evidence under the prescribed program and protocol; unresolved searches do not supply answer labels. Appendix~\ref{app:construction-verification} details candidate validation and content review.

\subsection{Benchmark Statistics}
\label{sec:dataset-profile}
\bench{} contains 1,200 questions from 200 scenarios across 30 tasks: 400 team-plan decisions and 800 optimality-verification variants. The 200 short- and 200 long-horizon questions form 200 pairs sharing the initial state. The variants form 400 pairs, each containing one question with the optimum retained and one with it removed. Both types of pairs derive from the same 200 scenarios. The two evaluation dimensions are scored separately, and option rotations do not count as additional questions.

In 64 scenarios, the short- and long-horizon questions use identical candidate programs and retain the same optimal choice, supporting comparisons across goals and execution protocols. In 15 long-horizon questions, a feasible alternative finishes its prescribed stage sooner than the optimal candidate but has a higher total completion cost. These cases support analysis of choices where stage length and final cost favor different plans. Appendix~\ref{app:detailed-statistics} reports structural groups for related recipe variants and programs, with grouping rules and full counts.

\section{Experiments and Analysis}
\label{sec:evaluation}
\label{sec:results}

\begin{table*}[!tp]
\centering\small
\renewcommand{\arraystretch}{1.04}
\definecolor{ResultTop1}{RGB}{177,207,232}
\definecolor{ResultTop2}{RGB}{208,226,241}
\definecolor{ResultTop3}{RGB}{234,243,250}
\begin{tabularx}{\textwidth}{@{}r>{\raggedright\arraybackslash}X>{\centering\arraybackslash}p{.13\textwidth}>{\centering\arraybackslash}p{.13\textwidth}>{\centering\arraybackslash}p{.19\textwidth}@{}}
\toprule
\textbf{Rank} & \textbf{Model} & \textbf{Params. (B)} & \textbf{DA} $\uparrow$ & \textbf{VA} $\uparrow$ \\
\midrule
\rowcolor{ResultTop1}
1 & \hfmodelhref{https://huggingface.co/PelaAI/KnowLine-31B-Gen1}{KnowLine-31B-Gen1} & 31 & \textbf{62.50} & 51.44 \\
\rowcolor{ResultTop2}
2 & \hfmodelhref{https://huggingface.co/Remek/basal-omni-1.0-ultra}{Basal-Omni-1.0-Ultra} & 31 & 60.75 & 51.97 \\
\rowcolor{ResultTop3}
3 & \hfmodelhref{https://huggingface.co/chinhnc/Quyet-1.0-Large}{Quyet-1.0-Large} & 31 & 60.75 & 52.06 \\
4 & \hfmodelhref{https://huggingface.co/dopest-studio/dopest-1.0-31b}{DOPEST-1.0-31B} & 31 & 60.38 & 52.16 \\
5 & \hfmodelhref{https://huggingface.co/perplexity-ai/pplx-decider-v1.2}{PPLX-Decider-v1.2} & 27 & 60.38 & 55.31 \\
6 & \hfmodelhref{https://huggingface.co/perplexity-ai/pplx-decider-v1.1-27b}{PPLX-Decider-v1.1-27B} & 27 & 59.44 & 54.28 \\
7 & \hfmodelhref{https://huggingface.co/torchcast-ai/torchcast-decision-4b}{Torchcast-4B-v1.2} & 4 & 58.94 & 50.66 \\
8 & \hfmodelhref{https://huggingface.co/h2oai/h2o-lightning-31b}{H2O-Lightning-31B} (BF16) & 31 & 58.69 & 50.19 \\
9 & \hfmodelhref{https://huggingface.co/vllm-sr/d3}{D3} & 27 & 58.69 & 52.19 \\
10 & \hfmodelhref{https://huggingface.co/fastino/GLiDE}{GLiDE-1.2-Open} & 27 & 58.13 & 54.72 \\
11 & \hfmodelhref{https://huggingface.co/startlux-models/StartLux-Decision-v1.1-27B-Instant}{StartLux-v1.1-27B-Instant} & 27 & 57.44 & 54.00 \\
12 & \hfmodelhref{https://huggingface.co/salfatigroup/rene-1-31b-fp8}{René-31B} (dequant. BF16) & 31 & 55.56 & 53.41 \\
13 & \hfmodelhref{https://huggingface.co/perplexity-ai/pplx-decider-flash-v1.2}{PPLX-Decider-Flash-v1.2} & 4 & 52.94 & 53.19 \\
14 & \hfmodelhref{https://huggingface.co/vllm-sr/d3-turbo}{D3-Turbo} & 27 & 52.94 & 49.97 \\
15 & \hfmodelhref{https://huggingface.co/torchcast-ai/torchcast-decision-4b}{Torchcast-4B-v1.3} & 4 & 52.75 & 51.28 \\
16 & \hfmodelhref{https://huggingface.co/google/gemma-4-31B-it}{deck31b} (BF16) & 31 & 52.19 & 49.72 \\
17 & \hfmodelhref{https://huggingface.co/org2ai/Wald-26B-A4B}{Wald-26B-A4B} & 26/4 & 50.63 & 51.09 \\
18 & \hfmodelhref{https://huggingface.co/jgeuter/Jeff-1.1-Medium}{Jeff-1.1-Medium} & 12 & 50.50 & 50.69 \\
19 & \hfmodelhref{https://huggingface.co/vllm-sr/Decision-2.0-Vega-27B}{Vega-27B} & 27 & 50.25 & 52.28 \\
20 & \hfmodelhref{https://huggingface.co/collections/morriszjm/tacit}{Tacit-9B} & 9 & 48.63 & 49.94 \\
21 & \hfmodelhref{https://huggingface.co/vllm-sr/d3-ultra}{D3-Ultra} & 27 & 47.50 & \textbf{56.28} \\
22 & \hfmodelhref{https://huggingface.co/MeTask-Rain/metask-jev-rain-31B-trained}{MeTask-Rain-31B} & 31 & 47.44 & 50.56 \\
23 & \hfmodelhref{https://huggingface.co/MeTask-Rain/metask-jev-rain-12B-trained}{MeTask-Rain-12B} & 12 & 47.38 & 50.66 \\
24 & \hfmodelhref{https://huggingface.co/LiquidAI/d1-3B}{Liquid AI d1-3B} & 3.12 & 47.19 & 53.47 \\
25 & \hfmodelhref{https://huggingface.co/torchcast-ai/torchcast-decision-12b}{Torchcast-12B} & 12 & 46.75 & 52.69 \\
26 & \hfmodelhref{https://huggingface.co/collections/morriszjm/tacit}{Tacit-4B} & 4 & 46.31 & 50.00 \\
27 & \hfmodelhref{https://huggingface.co/Mapika/decider-12b}{Mapika-Decider-12B} & 12 & 44.25 & 50.78 \\
28 & \hfmodelhref{https://huggingface.co/baselabs/mammoth-1}{Mammoth-1} & 27 & 43.88 & 52.34 \\
29 & \hfmodelhref{https://huggingface.co/torchcast-ai/torchcast-decision-4b}{Torchcast-4B-v1.1} & 4 & 43.50 & 52.34 \\
30 & \hfmodelhref{https://huggingface.co/Cloudflare/clef}{Clef} & 27 & 40.94 & 52.91 \\
31 & \hfmodelhref{https://huggingface.co/baselabs/joey-1}{Joey-1} & 2 & 40.69 & 50.53 \\
32 & \hfmodelhref{https://huggingface.co/collections/morriszjm/tacit}{Tacit-2B} & 2 & 39.44 & 50.25 \\
33 & \hfmodelhref{https://huggingface.co/h2oai/h2o-lightning-4b}{H2O-Lightning-4B} & 4 & 39.06 & 51.19 \\
34 & \hfmodelhref{https://huggingface.co/vllm-sr/Decision-2.0-Nox-4B}{Nox-4B} & 4 & 39.00 & 50.00 \\
35 & \hfmodelhref{https://huggingface.co/Cloudflare/clef-flash}{Clef-Flash} & 9 & 35.81 & 49.91 \\
36 & \hfmodelhref{https://huggingface.co/vllm-sr/Decision-2.0-Lux-9B}{Lux-9B} & 9 & 34.56 & 51.97 \\
37 & \hfmodelhref{https://huggingface.co/collections/morriszjm/tacit}{Tacit-1.7B} & 1.7 & 29.44 & 49.59 \\
38 & \hfmodelhref{https://huggingface.co/convaiinnovations/laya}{Laya-421M} & 0.421 & 25.88 & 49.69 \\
39 & \hfmodelhref{https://huggingface.co/vllm-sr/Decision-2.0-Kai-0.6B}{Kai-0.6B} & 0.6 & 21.81 & 50.00 \\
40 & \hfmodelhref{https://huggingface.co/Contrastive-LM/CLM-v0.1-8B}{CLM-8B} & 8 & 15.44 & 45.94 \\
\bottomrule
\end{tabularx}
\caption{DA and VA (\%) for 40 native decision models, averaged over four option rotations. Parameters are in billions; ``26/4'' denotes total/active parameters.}
\label{tab:results}
\end{table*}

\begin{table*}[!tp]
\centering\small
\renewcommand{\arraystretch}{1.04}
\definecolor{ResultTop1}{RGB}{239,196,151}
\definecolor{ResultTop2}{RGB}{247,219,189}
\definecolor{ResultTop3}{RGB}{252,239,225}
\begin{tabularx}{\textwidth}{@{}r>{\raggedright\arraybackslash}X>{\centering\arraybackslash}p{.19\textwidth}>{\centering\arraybackslash}p{.13\textwidth}>{\centering\arraybackslash}p{.13\textwidth}@{}}
\toprule
\textbf{Rank} & \textbf{Model} & \textbf{Avg. output tokens} & \textbf{DA} $\uparrow$ & \textbf{VA} $\uparrow$ \\
\midrule
\rowcolor{ResultTop1}
1 & \hfmodelhref{https://huggingface.co/Qwen/Qwen3.8-27B}{Qwen3.8-27B} & 8652 & \textbf{82.80} & \textbf{71.88} \\
\rowcolor{ResultTop2}
2 & \hfmodelhref{https://huggingface.co/google/gemma-4-31B-it}{Gemma4-31B} & 1072 & 73.11 & 57.03 \\
\rowcolor{ResultTop3}
3 & \hfmodelhref{https://huggingface.co/google/gemma-4-26B-A4B-it}{Gemma4-26B-A4B} & 1516 & 60.58 & 53.66 \\
4 & \hfmodelhref{https://huggingface.co/meta-models/Muse-Glimmer-30B}{Muse-Glimmer-30B} & 1415 & 59.64 & 60.35 \\
5 & \hfmodelhref{https://huggingface.co/Qwen/Qwen3-32B}{Qwen3-32B} & 925 & 46.33 & 49.45 \\
\bottomrule
\end{tabularx}
\caption{DA and VA (\%) and mean output length for the five highest-DA general-purpose models.}
\label{tab:general-results}
\end{table*}

\begin{figure*}[!t]
\centering
\includegraphics[width=\textwidth]{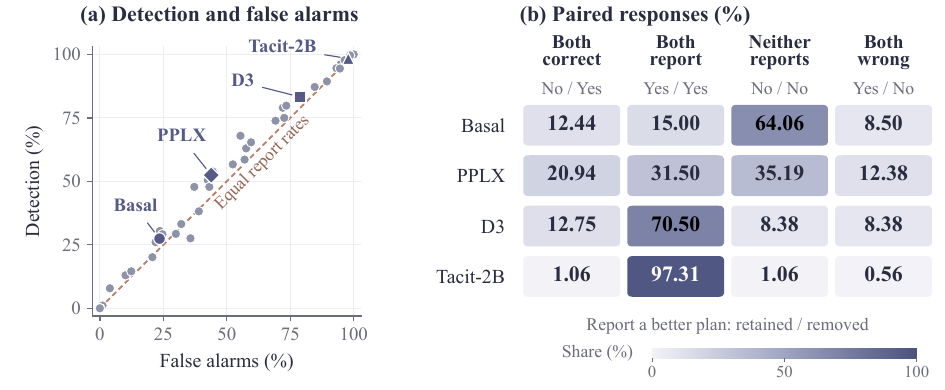}
\caption{Optimality-verification behavior. (a) Detection versus false-alarm rates. (b) Paired responses when the optimal plan is retained or removed, in that order. \emph{Yes} reports a better plan; \emph{No} selects a concrete plan.}
\label{fig:judgment-analysis}
\end{figure*}

\subsection{Experimental Setup}
\bench{} uses English inputs. We evaluate 40 native decision models and report reference results from 10 general-purpose language models. Unresolved answers from these models count as incorrect under the scoring policy. Each general-purpose model contributes 19,200 outputs across questions, option rotations, and samples.

Each question is evaluated under four cyclic option rotations, so each candidate appears once in each position. Native decision models produce one response per rotation, and general-purpose models produce four. Each response is scored separately, without voting. Models use their native decision or generation interfaces; Appendix~\ref{app:models} details reasoning modes and sampling parameters.

Decision accuracy (DA) measures selection of the optimal concrete plan. Verification accuracy (VA) averages the correct-detection rate when the optimum is removed and the correct-rejection rate when it is retained. Any valid concrete-plan choice counts as rejecting the better-plan statement. Invalid or truncated responses count as incorrect for both metrics. Uniform option selection yields 25\% DA and 50\% VA. Appendix~\ref{app:extended-evaluation} gives the metric formulas.

\subsection{Overall Performance}
Native decision models vary widely in team-plan decision accuracy, from 15.44\% to 62.50\% (Table~\ref{tab:results}). Their verification accuracy remains near chance, ranging from 45.94\% to 56.28\%.

The highest-DA model, KnowLine, does not lead in verification: D3-Ultra achieves the highest VA despite lower DA. Thus, plan-selection accuracy alone does not capture how well models judge whether better arrangements exist beyond the supplied options.

Among general-purpose models (Table~\ref{tab:general-results}; Appendix~\ref{app:model-comparisons}), Qwen3.8-27B leads with 82.80\% DA and 71.88\% VA.

\subsection{Optimality Verification}
\label{sec:judgment-analysis}
Many models report a better plan at similar rates whether the optimal plan is present or absent (Figure~\ref{fig:judgment-analysis}(a)). Tacit-2B detects 98.38\% of cases where the optimum is absent, but also raises false alarms in 97.88\% of cases where it is present.

Paired responses show two recurring patterns: some models report a better plan in both variants, while others select a concrete plan in both (Figure~\ref{fig:judgment-analysis}(b)). Both patterns produce a correct judgment in one condition and an incorrect judgment in the other.

The better-plan probabilities also discriminate weakly between the two conditions, with a median AUROC of 0.537 and a maximum of 0.598 across native models (Appendix~\ref{app:probability-details}).

\begin{figure}[!t]
\centering
\includegraphics[width=\columnwidth]{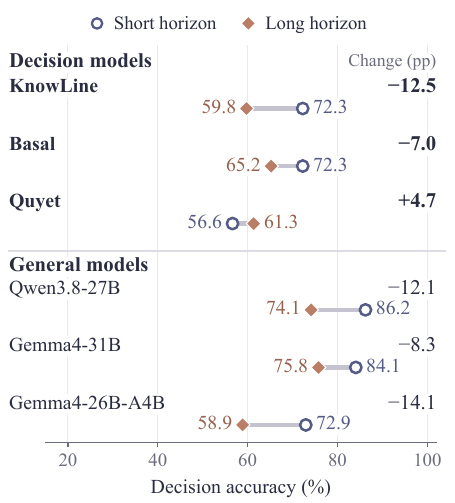}
\caption{Short- and long-horizon DA with identical candidate programs but different goals and execution rules. The three models with the highest overall DA in each family are shown. Changes are long minus short, in percentage points.}
\label{fig:goal-horizon-analysis}
\end{figure}

\begin{figure*}[!t]
\centering
\includegraphics[width=\textwidth]{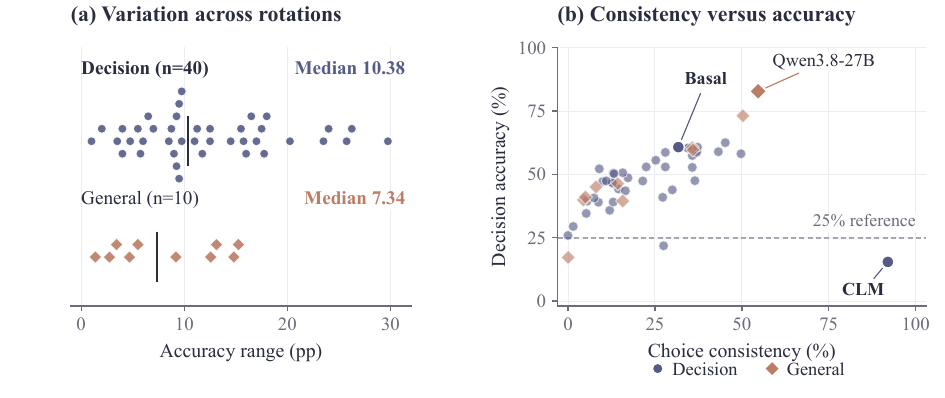}
\caption{Sensitivity to option order. (a) DA range across four rotations; bars mark group medians. (b) DA versus choice consistency, the percentage of questions selecting the same underlying plan across all rotations. General-model scores average four samples per rotation and include sampling variability.}
\label{fig:position-analysis}
\end{figure*}

\subsection{Short- and Long-Horizon Goals}
\label{sec:goal-analysis}
We compare 64 matched pairs that share the initial state, candidate programs, and optimal choice, but differ in their goal and execution rules.

Across native decision models, mean DA decreases from 49.58\% for short-horizon goals to 46.75\% for long-horizon goals, although individual models vary (Figure~\ref{fig:goal-horizon-analysis}). Full model-level results appear in Appendix~\ref{app:full-goal-analysis}.

Earlier stage completion can leave more work for the team. Appendix~\ref{app:analysis-details} reports model-level choices on 15 long-horizon questions where an alternative finishes its prescribed stage sooner but requires more joint steps to reach the final goal.

\subsection{Position Sensitivity}
\label{sec:position-analysis}
We measure position sensitivity as the difference between a model's highest and lowest accuracy across four option rotations. The median range across native models is 10.38 percentage points, indicating that option order affects measured accuracy (\mbox{Figure~\ref{fig:position-analysis}(a)}).

Choice consistency measures how often the same underlying plan is selected across all four rotations (\mbox{Figure~\ref{fig:position-analysis}(b)}). CLM selects the same plan in 92.00\% of questions despite a DA of 15.44\%. Stable choices can therefore still be consistently incorrect.

\section{Conclusion}
\label{sec:conclusion}
\bench{} evaluates decision models on team plans involving task dependencies, resource handoffs, and parallel work. Execution and exact search establish reference answers, while paired candidate sets test whether models recognize when better arrangements exist beyond the supplied options. Native decision models differ widely in plan-choice accuracy, but their reports of better alternatives often fail to distinguish whether an optimal plan is present. Performance also varies across goal settings and option orders. \bench{} therefore supports comparisons that assess both the selected team plan and judgments about the available alternatives.

\section*{Limitations}
Our findings concern team-plan decisions for two agents in one kitchen with global information. Questions share scenarios, and inference budgets vary across models.

\FloatBarrier
\bibliography{references}
\clearpage
\appendix
\twocolumn[\section*{Appendix}]
\section{Execution and Decision Criteria}
\label{app:semantics}
\subsection{Execution Rules}
Each joint step resolves chef interaction, assistant interaction, movement, then appliance processing. Processing includes its start step; finished products await retrieval. Waits are checked at step boundaries. An assistant interaction can retrieve an item placed by the chef in the same step; an explicit wait cannot.

Domains define role permissions, retrieval, and counter access. The canonical processing-chain domain permits retrieval before processing starts and plated-product transfers. Search records require matching states, goals, domains, and execution rules.

\subsection{Goal-Dependent Costs}
\label{app:formal-decisions}
Let $s_t^p$ be the state after $t$ joint steps of plan $p=(p^{(1)},p^{(2)})$, and $V_g^*(s;\mathcal D)$ the minimum legal cost to reach goal $g$ in domain $\mathcal D$. Short-horizon cost stops at the intermediate goal:
\begin{equation}
C_{\mathrm{S}}(s,p)=\min\{t\geq0:g_{\mathrm{S}}(s_t^p)=1\}.
\label{eq:short}
\end{equation}
If both programs finish at step $\tau_p$ before final delivery, long-horizon cost includes optimal continuation:
\begin{equation}
C_{\mathrm{L}}(s,p)=\tau_p+V_{g_{\mathrm{L}}}^*(s_{\tau_p}^p;\mathcal{D}).
\label{eq:long}
\end{equation}
Delivery during the prescribed stage instead uses its first hitting time. Members remain idle after their programs end, while appliances advance each step. A plan that cannot complete under this protocol has $C_g=+\infty$, even if abandoning it would permit recovery.

For the four base candidates, $i^*=\operatorname*{arg\,min}_{i\in\{1,2,3,4\}}C_g(s,p_i)$ is unique and satisfies
\begin{equation}
C_g(s,p_{i^*})=V_g^*(s;\mathcal{D}).
\label{eq:stateopt}
\end{equation}
State optimality ranges over all legal joint executions, with uniqueness required among the supplied candidates. For three retained plans $\mathcal P'$, the better-plan statement is true exactly when
\begin{equation}
\min_{p\in\mathcal{P}'} C_g(s,p)>V_g^*(s;\mathcal{D}).
\label{eq:missing-optimum}
\end{equation}

\subsection{Concrete Example}
\label{app:example}
In the long-horizon \texttt{point024} example (taro--bean--bell-pepper pie), B attains the 13-step optimum; C ends its stage sooner but requires 15 steps overall (Table~\ref{tab:concrete-example}). Inputs contain full programs without costs; Figure~\ref{fig:example} is schematic.

\begin{table}[!htb]
\centering\small
\begin{tabularx}{\columnwidth}{@{}lXrrr@{}}
\toprule
Plan & Program ID & Stage & Remaining & Total \\
\midrule
A & \texttt{opt04} & 13 & 8 & 21 \\
B & \texttt{opt01} & 10 & 3 & \textbf{13} \\
C & \texttt{opt02} & 7 & 8 & 15 \\
D & \texttt{opt03} & 9 & 6 & 15 \\
\bottomrule
\end{tabularx}
\caption{Long-horizon plan costs in joint steps. Stage is the prescribed-program cost; Remaining is the optimal continuation cost. Bold marks the minimum total.}
\label{tab:concrete-example}
\end{table}

\section{Construction and Dataset Details}
\label{app:construction-details}
\label{app:construction-procedures}
\subsection{Scenario Screening}
\label{app:screening-criteria}
Algorithm~\ref{alg:scenario-selection} selects states where coordination choices affect completion cost even when all remaining actions are optimized. A constraint $c$ fixes the order of two required events by one member or the location of a required handoff. Alternatives for the same events form a comparison family. With all remaining actions optimized, their cost penalty is
\begin{equation}
\Delta(s,c)=J(s,c)-V_{g_{\mathrm{L}}}^*(s;\mathcal{D}).
\label{eq:selection-loss}
\end{equation}
where $J(s,c)$ is the constrained optimal delivery cost. An examined family must contain two proven feasible choices satisfying
\begin{equation}
\Delta(s,c^+)=0,\qquad \Delta(s,c^-)\geq3.
\label{eq:selection-gate}
\end{equation}

\begin{algorithm}[t]
\caption{Scenario Selection}
\label{alg:scenario-selection}
\small
\begin{algorithmic}[1]
\Require Reachable states $S$ with domains $\{\mathcal D_s\}$;
\Statex \hspace{\algorithmicindent} trajectory lengths $\{T_i\}$, budget $B=200$;
\Statex \hspace{\algorithmicindent} comparison-family limit $K=8$
\Ensure Selected scenarios $Q$
\State $R\gets\emptyset$
\For{$s\in S$}
    \State $v_s\gets V_{g_{\mathrm L}}^*(s;\mathcal D_s)$
    \State $\mathcal F_s\gets\Call{ComparisonFamilies}{s,K}$
    \For{$F\in\mathcal F_s$}
        \State $J(s,c)\gets$ optimal delivery cost subject to $c$, for $c\in F$
        \State $P_F\gets$ choices in $F$ with proven finite costs
        \State $\Delta(s,c)\gets J(s,c)-v_s$ for $c\in P_F$
        \If{$\exists c^+,c^-\in P_F$ satisfying Eq.~\ref{eq:selection-gate}}
            \State $R\gets R\cup\{s\}$
        \EndIf
    \EndFor
\EndFor
\State $(G_i)\gets\Call{GroupComparisons}{R}$
\State $(q_i)\gets\Call{AllocateQuotas}{(G_i),\{T_i\},B}$
\State $Q\gets\emptyset$
\For{each task $i$}
    \State $Q\gets Q\cup\Call{GreedyCoverage}{G_i,q_i}$
\EndFor
\State \Return $Q$
\end{algorithmic}
\end{algorithm}

States match by positions, orientations, carried items, material locations, and processing progress. Up to eight comparison families alternate member-specific event orders and both handoff directions at dependency depths 1--3. Alternatives are fixed before cost queries.

Grouping preserves task, domain, goal, comparison family, alternatives, feasibility, penalties, and timing relative to partner events. Materials and processing progress remain fixed across intervening trajectory states; positions may vary. The representative is nearest a certified divergence, with earlier trajectory position breaking ties.

\subsection{Coverage and Candidate Verification}
\label{app:construction-verification}
For representative states $S_i$ and optimal trajectory length $T_i$, task capacity is
\begin{equation}
b_i=\min\bigl(|S_i|,\lfloor T_i/3\rfloor\bigr).
\label{eq:scenario-capacity}
\end{equation}
Capacity limits counts, not spacing. Each task receives one scenario; the remaining 170 follow largest-remainder allocation proportional to $b_i-1$, with ties by task ID. Coverage prioritizes event-ordering or handoff directions, family/optimal-arrangement-set combinations, then material stages. Keys include all optimal arrangements; processing progress alone does not define a stage.

Ties use trajectory distance, proximity grade, earlier position, then candidate ID. Grades order certified common optimal prefixes of 0, 1, 2, or $\geq3$ steps, then unknown and uncomputed cases. Source levels L1--L6 contribute 10, 5, 40, 35, 50, and 60 scenarios; these are recipe-complexity labels, not calibrated item difficulties.

Base questions contain a unique state-optimal plan and at least two feasible alternatives. Suboptimal gaps are at least one joint step for short goals and two for long goals; short-goal contrasts act before first attainment. Rationales link state facts and program differences to execution outcomes. Parallelism tags require different productive overlap.

\subsection{Dataset Composition}
\label{app:detailed-statistics}
Table~\ref{tab:profile-detailed} separates questions, scenarios, and structures. Of 200 cross-goal pairs, 64 share programs and the optimum; 136 use different programs. Normalization replaces ingredient/product names while preserving dependencies, states, programs, goals, and domains.
\begin{table}[!htb]
\centering\small
\begin{tabularx}{\columnwidth}{@{}Xr@{}}
\toprule
Unit or property & Count \\
\midrule
Recipe tasks & 30 \\
Decision scenarios & 200 \\
Short-/long-horizon question pairs & 200 \\
Total evaluation questions & 1,200 \\
Base team-plan decision questions & 400 \\
Optimality-verification variants & 800 \\
Base short-/long-horizon questions & 200 / 200 \\
Base candidate plans & 1,600 \\
Base questions with 3 / 4 feasible plans & 246 / 154 \\
Normalized question structures & 181 \\
Normalized paired-scenario structures & 97 \\
Pairs with identical candidate programs & 64 \\
Long-horizon questions with stage/total rank reversals & 15 \\
\bottomrule
\end{tabularx}
\caption{Dataset composition. Candidate and feasibility counts cover base questions; normalized structures cover base questions and short-/long-horizon pairs. Option rotations do not create additional questions.}
\label{tab:profile-detailed}
\end{table}

The public dataset release is \texttt{tandembench-en-v1.0.0}. Original gold positions A--D contain 200, 69, 62, and 69 questions; cyclic rotations balance positions.

\section{Models and Inference Protocols}
\label{app:models}
\paragraph{Native interfaces.}
Models retain native inputs and readouts: Tacit inference; Decision-2 for Vega/Lux/Kai/Nox; CLM's prediction head and last-token pooling; Laya state inputs and short labels; KnowLine's \texttt{read\_twice}; and GLiDE's non-causal text interface. Mammoth, PPLX, and D3 retain calibration; StartLux uses Instant inference. MeTask-Rain-31B uses W4A16 with unmerged LoRA, routing, and calibration.

\paragraph{Weights and reasoning.}
H2O-Lightning-31B and deck31b use BF16; Ren\'e uses FP8 dequantized to BF16. H2O disables online FP8 quantization. D3 sizes round 26.09B to 27B; Wald lists total/active parameters. Muse-Glimmer-30B uses low reasoning, and the evaluated Qwen conditions disable thinking.

\paragraph{Response scoring.}
Truncated, missing-choice, and multiple-choice outputs count as incorrect.

\section{Evaluation Metrics and Full Results}
\label{app:extended-evaluation}
\subsection{Coverage and Scoring}
\begin{table*}[!tp]
\centering\small
\renewcommand{\arraystretch}{1.04}
\definecolor{ResultTop1}{RGB}{239,196,151}
\definecolor{ResultTop2}{RGB}{247,219,189}
\definecolor{ResultTop3}{RGB}{252,239,225}
\begin{tabularx}{\textwidth}{@{}r>{\raggedright\arraybackslash}X>{\centering\arraybackslash}p{.19\textwidth}>{\centering\arraybackslash}p{.13\textwidth}>{\centering\arraybackslash}p{.13\textwidth}@{}}
\toprule
\textbf{Rank} & \textbf{Model} & \textbf{Avg. output tokens} & \textbf{DA} $\uparrow$ & \textbf{VA} $\uparrow$ \\
\midrule
\rowcolor{ResultTop1}
1 & \hfmodelhref{https://huggingface.co/Qwen/Qwen3.8-27B}{Qwen3.8-27B} & 8652 & \textbf{82.80} & \textbf{71.88} \\
\rowcolor{ResultTop2}
2 & \hfmodelhref{https://huggingface.co/google/gemma-4-31B-it}{Gemma4-31B} & 1072 & 73.11 & 57.03 \\
\rowcolor{ResultTop3}
3 & \hfmodelhref{https://huggingface.co/google/gemma-4-26B-A4B-it}{Gemma4-26B-A4B} & 1516 & 60.58 & 53.66 \\
4 & \hfmodelhref{https://huggingface.co/meta-models/Muse-Glimmer-30B}{Muse-Glimmer-30B} & 1415 & 59.64 & 60.35 \\
5 & \hfmodelhref{https://huggingface.co/Qwen/Qwen3-32B}{Qwen3-32B} & 925 & 46.33 & 49.45 \\
6 & \hfmodelhref{https://huggingface.co/google/gemma-4-E4B-it}{Gemma4-E4B} & 2564 & 45.03 & 51.16 \\
7 & \hfmodelhref{https://huggingface.co/Qwen/Qwen3-8B}{Qwen3-8B} & 1092 & 41.03 & 49.85 \\
8 & \hfmodelhref{https://huggingface.co/mistralai/Ministral-3-14B-Instruct-2512-BF16}{Ministral 3 14B Instruct} & 1762 & 39.89 & 51.45 \\
9 & \hfmodelhref{https://huggingface.co/meta-llama/Llama-3.3-70B-Instruct}{Llama3.3-70B} & 661 & 39.50 & 48.92 \\
10 & \hfmodelhref{https://huggingface.co/Qwen/Qwen3-0.6B}{Qwen3-0.6B} & 25 & 17.19 & 40.61 \\
\bottomrule
\end{tabularx}
\caption{DA and VA (\%) and mean output length for ten general-purpose models, ranked by DA. Unresolved answers count as incorrect; shading marks the top three and bold marks the best score in each metric.}
\label{tab:general-results-full}
\end{table*}

For $N$ questions, decision accuracy (denoted $\mathrm{Acc}_{4,S}$ below) compares stable candidate IDs $\hat i_{qrs}$ with gold $i_q^*$:
\begin{equation}
\mathrm{Acc}_{4,S} =
\frac{1}{4SN}\sum_{q=1}^{N}\sum_{r=0}^{3}\sum_{s=1}^{S}
\ind[\hat{i}_{qrs}=i_q^*].
\label{eq:acc}
\end{equation}
Each model contributes 1,600 native or 6,400 general-model base outputs. Uniform selection or a fixed display letter yields 25\% DA. Truncated, missing-choice, multiple-choice, and infrastructure-failure outputs count as incorrect and are recorded separately.

Verification uses 800 paired variants. Reporting a better plan is TP when the optimum is removed; selecting any valid concrete plan is TN when it is retained:
\begin{equation}
\mathrm{VA}=\frac{1}{2}\left(\frac{\mathrm{TP}}{N_{\mathrm{remove}}}+\frac{\mathrm{TN}}{N_{\mathrm{retain}}}\right).
\label{eq:oja}
\end{equation}
VA is balanced binary accuracy and, with equal group sizes, pooled judgment accuracy. Invalid responses count as neither TP nor TN. Uniform selection, always reporting, or always choosing a concrete plan yields 50\% VA. TN does not require selecting the optimal retained plan.

\subsection{Full Model Comparison}
\label{app:model-comparisons}
Table~\ref{tab:general-results-full} reports results for the ten general-purpose models evaluated on the English inputs.

\subsection{Interpretation Scope}
The benchmark evaluates centralized plan decisions from complete global information; it does not test communication or adaptation to independent partners. Option rotations balance answer positions, while program length and waiting patterns remain observable cues. Models use their native interfaces and inference budgets, so results compare evaluated systems under their respective protocols.

\section{Costs and Goal Comparisons}
\label{app:analysis-details}
\subsection{Decision Costs and Choice Stability}
\begin{figure*}[!tp]
\centering
\includegraphics[width=\textwidth]{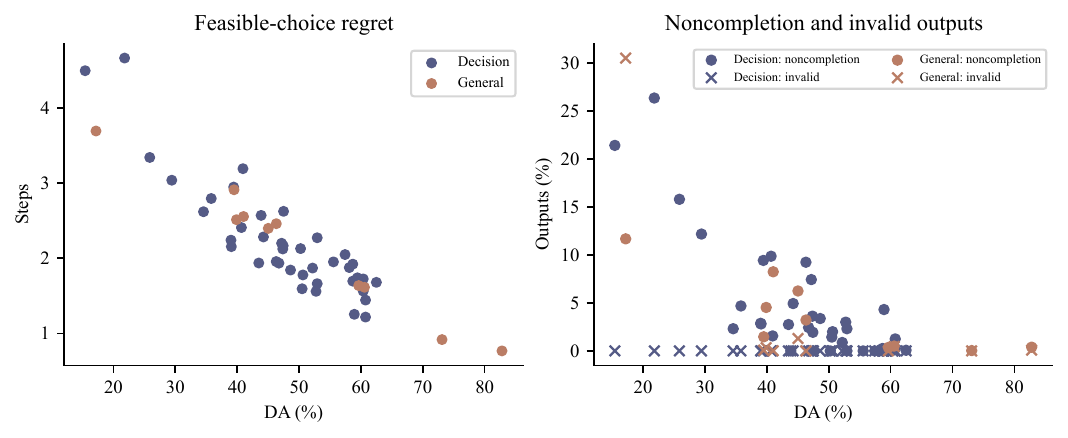}
\caption{Execution consequences of selected plans. Left: mean extra joint steps above the state optimum among feasible selections. Right: the percentage of all outputs selecting a noncompleting plan; invalid responses form a separate category.}
\label{fig:decision-costs}
\end{figure*}

Figure~\ref{fig:decision-costs} covers 40 native models and 10 general-purpose models; Table~\ref{tab:behavior} lists six native models. For feasible selections, regret is
\begin{equation}
R_q=C_{g_q}(s_q,p_{\hat{i}_q})-V_{g_q}^*(s_q;\mathcal{D}_q).
\label{eq:regret}
\end{equation}
Regret includes optimal choices at zero. Feasible, noncompletion, and invalid rates use all outputs and sum to one. The 246 questions with noncompleting candidates give exposure counts of 984 ($S=1$) or 3,936 ($S=4$).
\begin{table*}[t]
\centering\small

\begin{tabular}{@{}lrrrrr@{}}
\toprule
Model & Feasible (\%) & Noncompletion (\%) & Regret & Consistency (\%) & Range (pp) \\
\midrule
Basal-Omni-1.0-Ultra & 99.75 & 0.25 & 1.44 & 31.75 & 11.25 \\
KnowLine-31B-Gen1 & 99.94 & 0.06 & 1.68 & 45.25 & 11.75 \\
Quyet-1.0-Large & 98.75 & 1.25 & 1.22 & 37.25 & 25.75 \\
PPLX-Decider-v1.1-27B & 100.00 & 0.00 & 1.74 & 37.00 & 9.00 \\
Torchcast-4B-v1.2 & 95.69 & 4.31 & 1.25 & 43.25 & 2.00 \\
CLM-8B & 78.56 & 21.44 & 4.50 & 92.00 & 1.00 \\
\bottomrule
\end{tabular}
\caption{Execution outcomes and choice stability. Regret is mean extra joint steps among feasible selections. Consistency is the percentage of questions selecting the same plan across all rotations; Range is maximum minus minimum rotation DA.}
\label{tab:behavior}
\end{table*}

\paragraph{Position sensitivity.}
\label{app:position-details}
For general-purpose models, averaging four samples per position yields a median DA range of 7.34 points. Averaging the four within-sample ranges for each model and taking the median across models gives 8.47 points. Consistency tracks the same candidate across rotations within each sample index. Both include generation randomness (Figure~\ref{fig:position-analysis}; Table~\ref{tab:behavior}).

\subsection{Goal Comparisons}
\label{app:full-goal-analysis}
\begin{figure*}[!tp]
\centering
\includegraphics[width=\textwidth,height=.8\textheight,keepaspectratio]{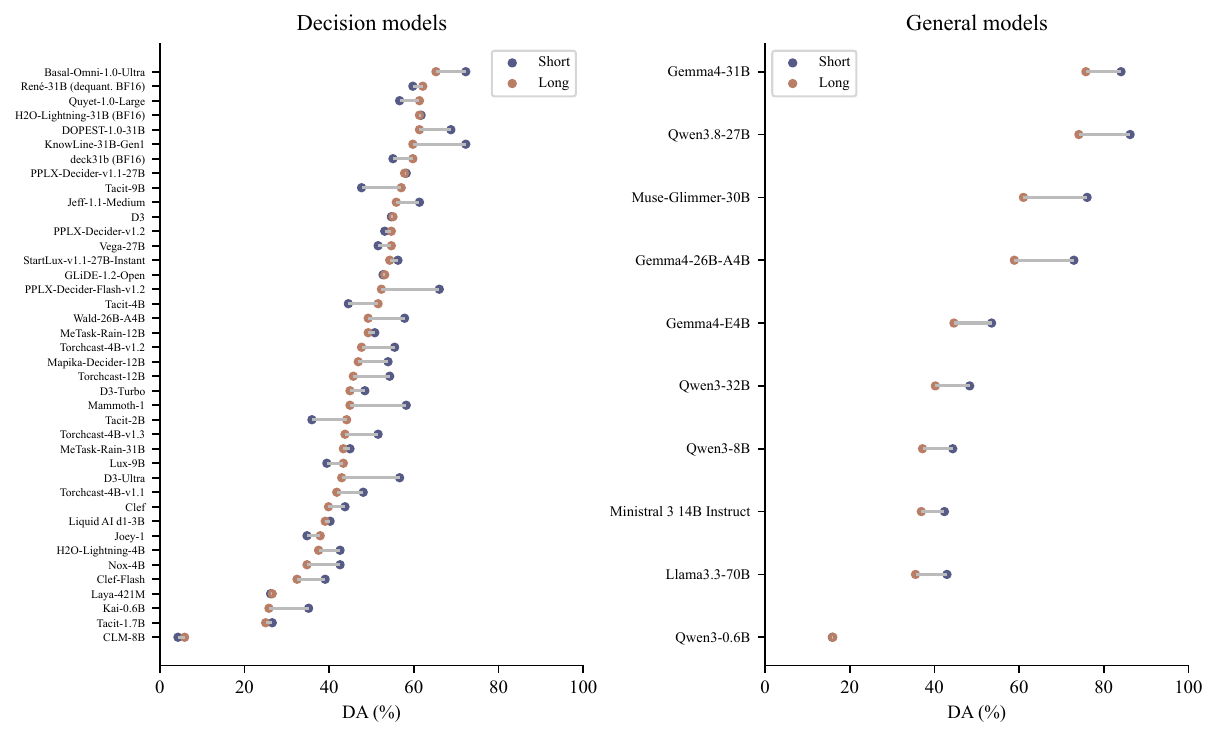}
\caption{Short- and long-horizon DA with identical candidate programs but different goals and execution rules. Models are ordered by decreasing long-horizon DA within each family.}
\label{fig:full-goal-analysis}
\end{figure*}

Figure~\ref{fig:full-goal-analysis} compares 64 pairs sharing initial states, programs, and optimal choices, with different goals and execution rules. Mean short/long DA is 49.58/46.75\% for 40 native models and 56.68/48.04\% for 10 general-purpose models, weighting models equally.
\begin{table*}[t]
\centering\small

\begin{tabular}{@{}lrrrr@{}}
\toprule
Model & Same-plan short & Same-plan long & Conflict DA & Fast-stage choice \\
\midrule
Basal-Omni-1.0-Ultra & 72.27 & 65.23 & 35.00 & 36.67 \\
KnowLine-31B-Gen1 & 72.27 & 59.77 & 55.00 & 0.00 \\
Quyet-1.0-Large & 56.64 & 61.33 & 48.33 & 10.00 \\
PPLX-Decider-v1.1-27B & 58.20 & 57.81 & 45.00 & 35.00 \\
Torchcast-4B-v1.2 & 55.47 & 47.66 & 73.33 & 15.00 \\
CLM-8B & 4.30 & 5.86 & 0.00 & 66.67 \\
\bottomrule
\end{tabular}
\caption{DA and fast-stage choice rates (\%). Same-plan columns compare short and long goals with identical programs. Conflict columns cover a separate set where an earlier stage finish can require more total steps; Fast-stage choice is the rate of selecting such alternatives.}
\label{tab:paired}
\end{table*}

A disjoint set of 15 long-horizon questions offers feasible alternatives with $\tau_i<\tau_{i^*}$ but $C_{\mathrm L}(s,p_i)>C_{\mathrm L}(s,p_{i^*})$. Table~\ref{tab:paired} reports their DA and selection rates. Basal selects these earlier-finishing but higher-total-cost alternatives in 22/60 presentations; Qwen3.8-27B in 13/240, including rotations and samples.

\section{Coordination-Group Analysis}
\label{app:mechanism-details}
\begin{figure*}[!tp]
\centering
\includegraphics[width=.8\textwidth]{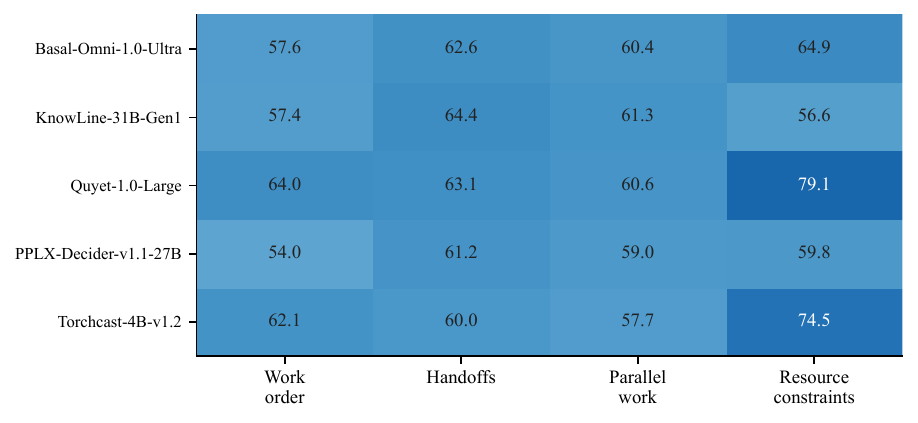}
\caption{DA (\%) across coordination groups. Darker cells indicate higher accuracy. A question may belong to more than one group.}
\label{fig:coordination-analysis}
\end{figure*}

Figure~\ref{fig:coordination-analysis} compares five native models across overlapping coordination groups defined by the union of candidate tags.

\begin{figure*}[!tp]
\centering
\includegraphics[width=\textwidth]{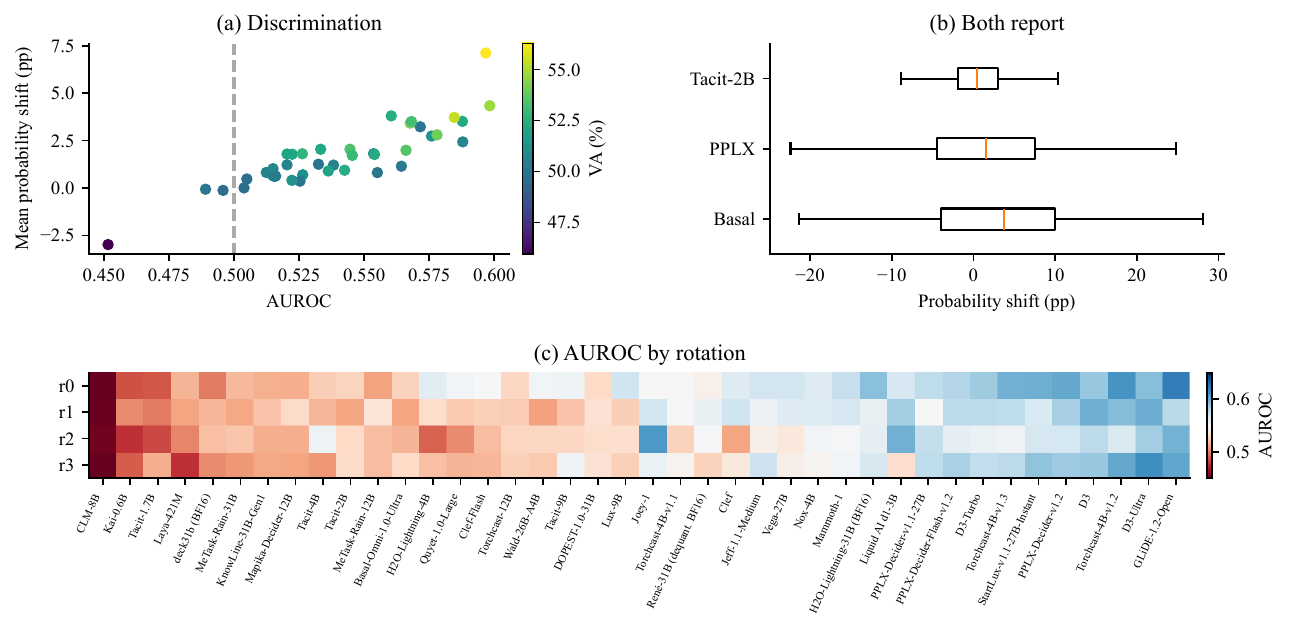}
\caption{Better-plan probabilities. (a) AUROC versus mean shift, colored by VA. (b) Shifts for pairs reporting twice; boxes show quartiles and central lines the median. (c) AUROC by rotation. Shifts are removed minus retained, in percentage points.}
\label{fig:probability-details}
\end{figure*}

Work order covers processing/supply order (300 questions, 156 scenarios); handoffs cover transfer locations/conditions (376, 188); parallel work covers productive overlap (355, 182); resource constraints cover carrying/counter capacity (141, 77). These tags capture intended coordination contrasts. Because groups may overlap and differ in task and goal composition, their scores describe performance on these subsets rather than isolated mechanism effects.

\section{Probability Analysis}
\label{app:probability-details}
Figure~\ref{fig:probability-details} uses 128,000 outputs: 40 models, 400 pairs, four rotations. Probabilities are aligned by option letter or text, correcting sums within $10^{-4}$ rounding tolerance and excluding interface-specific confidence fields.

AUROC ranks removed- against retained-optimum examples by better-plan probability, with half credit for ties. Median/maximum AUROC is 0.537/0.598 (GLiDE); Basal scores 0.520.

Panel (a) averages all pairs. For pairs reporting twice in (b), mean shifts are +3.26 points for Basal (240 pairs), +2.05 for PPLX (504), and +0.55 for Tacit-2B (1,557). Figure~\ref{fig:judgment-analysis}(b) gives unconditional choices.

\section{Task Weighting}

\paragraph{Task weighting and uncertainty.}
\begin{table}[!htb]
\centering\small
\setlength{\tabcolsep}{3pt}
\begin{tabular}{@{}lrr@{}}
\toprule
Model & Item-weighted & Task-macro \\
\midrule
Basal-Omni-1.0-Ultra & 60.75 & 65.45 \\
KnowLine-31B-Gen1 & 62.50 & 66.96 \\
Quyet-1.0-Large & 60.75 & 61.00 \\
PPLX-Decider-v1.1-27B & 59.44 & 62.90 \\
Torchcast-4B-v1.2 & 58.94 & 56.95 \\
CLM-8B & 15.44 & 10.52 \\
\bottomrule
\end{tabular}
\caption{DA (\%) under two weighting schemes. Item-weighted gives equal weight to questions; Task-macro averages task-level accuracies equally.}
\label{tab:task-weighting}
\end{table}

After averaging four rotations per question, we bootstrap 200 scenarios with both goals retained (10,000 replications; seed 20261027). KnowLine minus Basal is 1.75 points: the percentile 95\% interval is [-1.56, 5.00], or [-2.10, 5.56] when resampling the 30 tasks. Intervals condition on the recorded runs; structure IDs record further cross-scenario relationships.

\end{document}